\documentclass[journal]{IEEEtranTIE}
\usepackage{graphicx}
\usepackage{cite}
\usepackage{picinpar}
\usepackage{amsmath}
\usepackage{url}
\usepackage{flushend}
\usepackage[latin1]{inputenc}
\usepackage{colortbl}
\usepackage{soul}
\usepackage{multirow}
\usepackage{pifont}
\usepackage{color}
\usepackage{alltt}
\usepackage[hidelinks]{hyperref}
\usepackage{enumerate}
\usepackage{siunitx}
\usepackage{breakurl}
\usepackage{epstopdf}
\usepackage{pbox}
\usepackage{multicol}
\usepackage{multirow} 
\newcommand{\tabincell}[2]{\begin{tabular}{@{}#1@{}}#2\end{tabular}}
\usepackage{booktabs}

\begin{document}
\title{Improvement of Normal Estimation for Point Clouds via Simplifying Surface Fitting}

\author{
	\vskip 1em
	Jun Zhou$^*$,
	Wei Jin,
	Mingjie Wang,
	Xiuping Liu,
	Zhiyang Li,
	Zhaobin Liu
	\thanks{
		
		
	
	This work is supported by the National Natural Science Foundation of China (No. 62002040, No.61976040).
	
	J. Zhou, Z. Li, Z. Liu are with the School of information science and technology, Dalian Maritime University, China
	
	W. Jin is with Dalian Neusoft University of Information,China.
	
	M. Wang is with Memorial University of Newfoundland, Canada.
	
	X. Liu is with with Dalian University of Liaoning Province, China.
	
	E-mail: jun90@dlmu.edu.cn
	}
}

\maketitle
	
\begin{abstract}
With the burst development of neural networks in recent years, the task of normal estimation has once again become a concern. By introducing the neural networks to classic methods based on problem-specific knowledge, the adaptability of the normal estimation algorithm to noise and scale has been greatly improved. However, the compatibility between neural networks and the traditional methods has not been considered. Similar to the principle of Occam's razor, that is, the simpler is better. We observe that a more simplified process of surface fitting can significantly improve the accuracy of the normal estimation. In this paper, two simple-yet-effective strategies are proposed to address the compatibility between the neural networks and surface fitting process to improve normal estimation. Firstly, a dynamic top-k selection strategy is introduced to better focus on the most critical points of a given patch, and the points selected by our learning method tend to fit a surface by way of a simple tangent plane, which can dramatically improve the normal estimation results of patches with sharp corners or complex patterns. Then, we propose a point update strategy before local surface fitting, which smooths the sharp boundary of the patch to simplify the surface fitting process, significantly reducing the fitting distortion and improving the accuracy of the predicted point normal. The experiments analyze the effectiveness of our proposed strategies and demonstrate that our method achieves SOTA results with the advantage of higher estimation accuracy over most existed approaches.
\end{abstract}


\markboth{}
{}

\definecolor{limegreen}{rgb}{0.2, 0.8, 0.2}
\definecolor{forestgreen}{rgb}{0.13, 0.55, 0.13}
\definecolor{greenhtml}{rgb}{0.0, 0.5, 0.0}

\section{Introduction}

\IEEEPARstart{D}{ue} to the advances in 3D acquisition technologies, point-based representations are widely employed in various vision applications. However, the scanned unstructured point cloud frequently contains sampling irregularity while lacking the essential geometric property. As a fundamental feature, the surface normal can facilitate a considerable amount of downstream analyzing and processing works for the point cloud, such as 3D surface reconstruction\cite{berger2014state,kazhdan2006poisson,hashimoto2019normal,fleishman2005robust}, registration\cite{pomerleau2015review}.

It is well known that the classic normal estimation methods such as PCA-based method\cite{hoppe1992surface} or Jets\cite{cazals2005estimating} have been widely used in industry and achieved great success. This type of method takes into account the potential structural properties of the point cloud. It uses a well-designed and knowledge-related algorithm to fit the local tangent plane. However, these methods heavily rely on careful parameter tuning for handling the multiple types of noise, outliers, and sampling anisotropy.

\begin{figure}[!t]\centering
	\includegraphics[width=8.5cm]{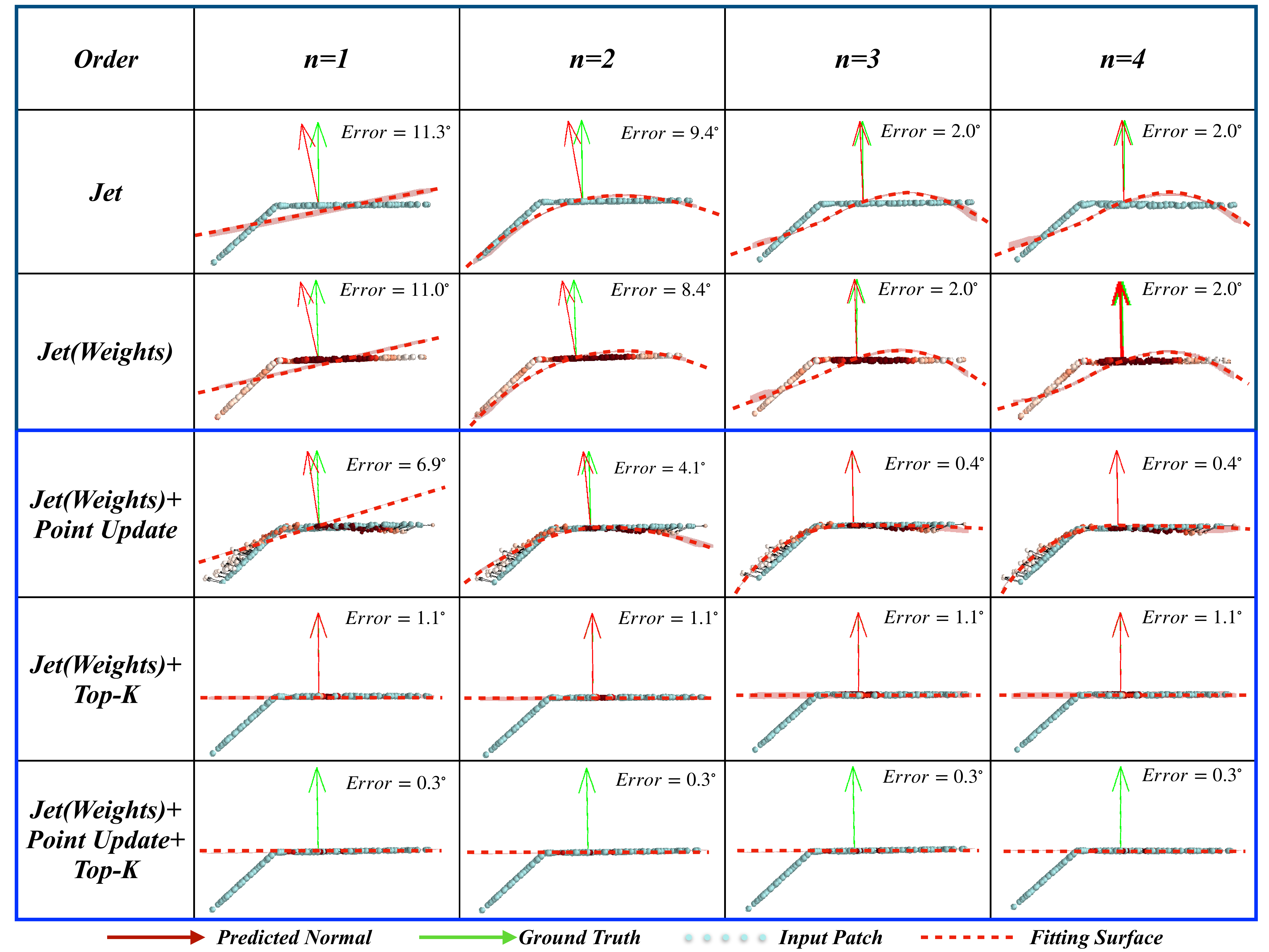}
	\caption{The 2D illustration of the surface fitting process for the classic Jet, Jet with weight learning neural network and our proposed simplifying strategies. With the surface fitting order increasing, our simplification strategies can be intended to explore a lower order to fit the local patch and improve normal estimation. The red dashed line indicates the 2D fitting surface, the green and red arrows indicate the ground-truth and predicted normal, respectively, and the value given on the upper right corner shows the angle error.}\label{FIG_1}
\end{figure}

\begin{figure*}[!t]\centering
	\includegraphics[width=17cm]{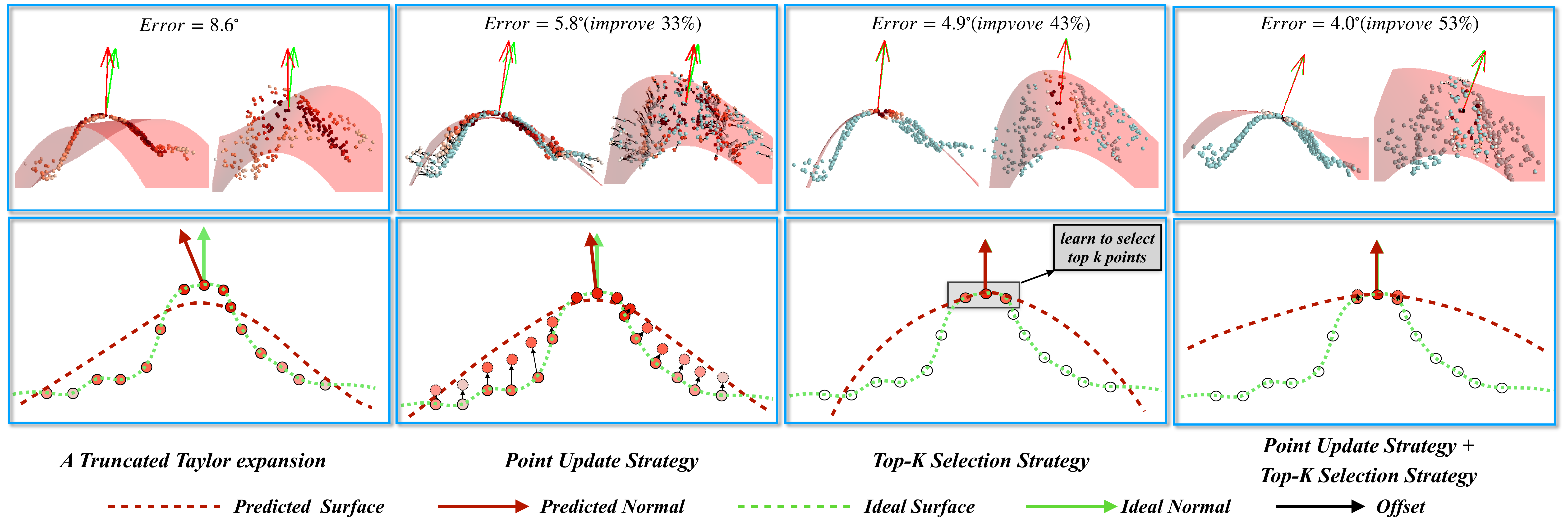}
	\caption{The visualizations of local surface fitting operations with different strategies. The top row: the 3D surface fitting process; The bottom row: the 2D illustrations of various operations. Compared to the traditional truncated Taylor expansion method, the point update strategy can push the points to construct a more smooth patch, and the proposed top-k selection strategy can dynamically choose the critical points of the given patch. The proposed strategies can simplify the fitting process and improve the accuracy of normal estimation for the point cloud.}\label{FIG_2}
\end{figure*}

The neural networks provide one solution for this dilemma since they can adapt to the statistics of the dataset. Most learned-based methods convert the normal estimation problem to a direct vector regression process and make remarkable progress. However, lack of problem-specific knowledge makes the regression network unstable to the point position's slight disturbance. Thus, MRTNet\cite{9339852} introduces a RANSAC-like module to approximate the local tangent space, and DeepFit\cite{ben2020deepfit} employs a Taylor expansion to more effectively describe the local point cloud surface. This kind of combination method has proven the improvement to the 
pure network regression. Still, there is a question: whether the combination of neural networks and traditional methods is fully compatible, whether there is a gap between them? As shown in the first two rows of Fig. \ref{FIG_1}, similar to DeepFit\cite{ben2020deepfit}, the truncated Taylor expansion (Jet) and a neural network are both used to estimate a point normal near a sharped boundary. However, it is disappointed that the weights predicted by the neural network cannot effectively improve the accuracy of surface fitting on patches with sharp features, even if increasing the n-jet orders (from order 1 to 4). We can observe two facts: 1) The generalization of the network will have a significant impact on weight learning, and the lack of practical focus will make the fitting process difficult; 2) The fitting process near sharp edges is complicated, which will affect the accuracy of point cloud normal estimation.

To address the compatibility problem between neural networks and the classic Jet method, we propose two simple and effective strategies, namely the dynamic top-k selection strategy and the point update strategy. First of all, the proposed top-k selection strategy can dynamically focus on the critical points while making the surface fitting process more straightforward. The 2D illustration can be shown in the third column of Fig \ref{FIG_2}. Then, the point update strategy is proposed to learn to adjust the positions of a given patch, and the updated points can be fitted more easily through a more concise parametric surface, which can greatly reduce the complexity for the Jet process (See the second column of Fig. \ref{FIG_2}). These two strategies can dramatically improve the compatibility between the classic Jet method and weight learning network. Besides, Fig. \ref{FIG_1} shows the comparison of the angular errors between classic Jet, DeepFit\cite{ben2020deepfit} and ours. Our proposed strategies can simplify the process of surface fitting and reduce the distortion of the fitting surface, which dramatically improves the accuracy of normal estimation results near the sharp corners.

The contributions of our work are summarized as follows:
\begin{itemize}
\item This paper proposes a point update strategy for the patch. The point update sub-network can effectively push the points of a given patch to better positions, making the surface fitting process easier to deal with, especially for a patch with sharp edges.

\item A dynamic selection mechanism for the top k weights is proposed, namely top-k selection strategy. This method ensures that a more precise weight learning is adopted for a given patch and achieves a more straightforward fitting process via focus the critical points.

\item Based on our proposed strategies, the combination between the weight-based learning network and the classic surface fitting method becomes more compatible, which shows great improvements over the state-of-the-arts.
\end{itemize}

\section{Related Work}
The task of normal estimation for point clouds is a long-standing basic problem in geometry processing, such as 3D shape reconstruction. In this section, we will review the traditional and popular deep-based methods for normal estimation. Firstly, we will give a brief description of the 3D point deep learning history.

\subsection{Deep learning for 3D point clouds}
The processing of unordered 3D point clouds faces several significant challenges, such as the unstructured nature and varying sampling density. The voxel-based methods\cite{maturana2015voxnet,wu20153d,qi2016volumetric} are the pioneers applying 3D convolutional neural networks, which directly extend 2D to 3D grids. Similarly, the 3D modified fisher vectors (3DmFV)\cite{ben20183dmfv} maps the point deviation from a Gaussian Mixture Model (GMM) to a coarse grid. Due to the accuracy-complexity trade-off, the Kd-Networks \cite{klokov2017escape,riegler2017octnet} are imposed on the points to learn shared weights for nodes in the tree, which can remarkably reduce memory footprints and improve computations. Recently, the PointNet\cite{qi2017pointnet,qi2017pointnet++} and its variants apply the symmetric, order-insensitive function on a high-dimensional representation of individual points, which can directly manipulate point clouds and have made remarkable progress in lots of areas of 3D vision.

\begin{figure*}[htp!]
 \centering 
 \includegraphics[width=16cm]{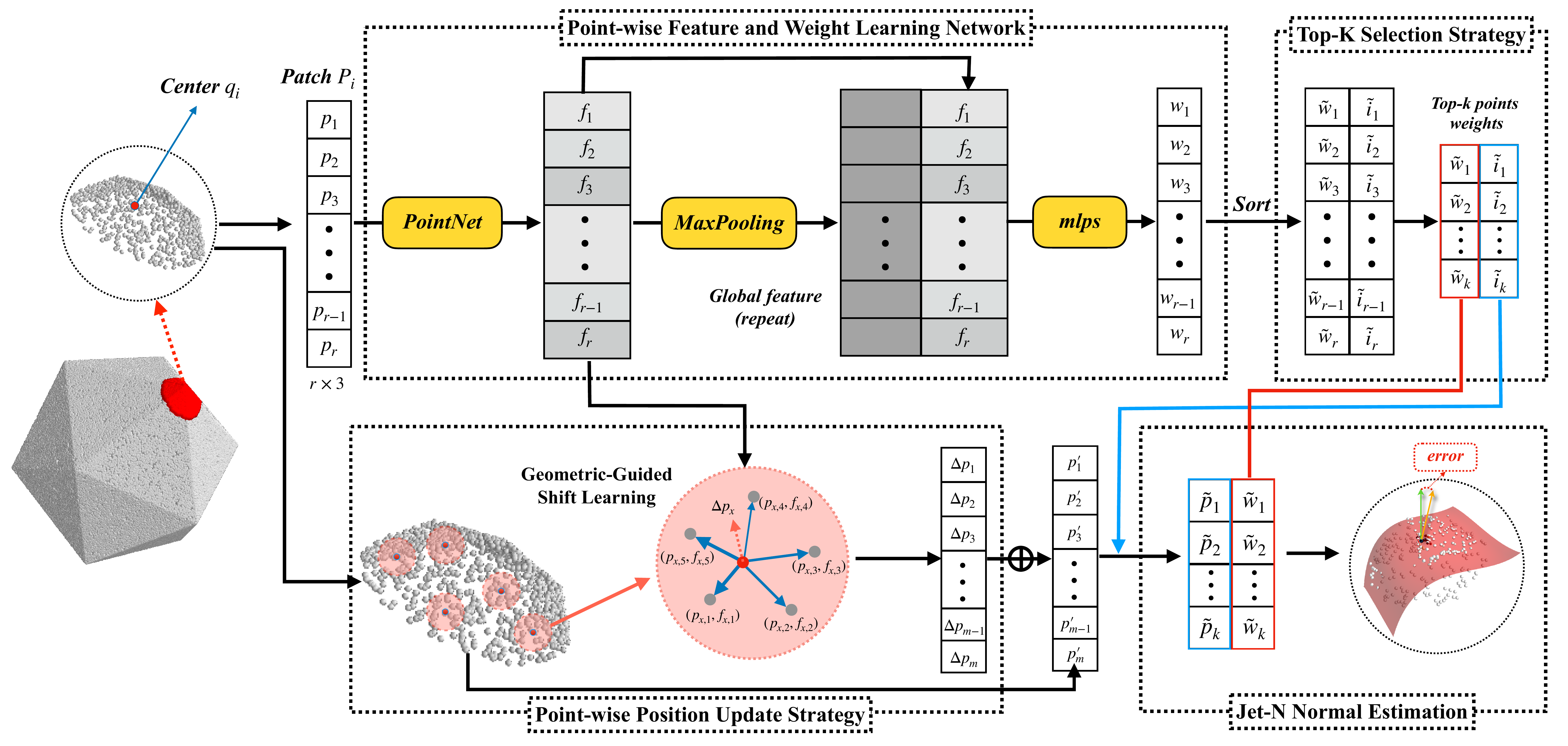}
 \caption{The pipeline of our normal estimation method. Given a point cloud $X$, we employ the trained network to estimate the center's normal from the sampled patches. The learned weights of the patch are used to select the critical points, and a Jet method is used to estimate the normal of the patch. The top-k selection strategy is trained in our network to choose the most significant points dynamically. The point update module is a sub-network for adaptively adjusting the given patch structure to simplify the Jet-based normal estimation.}
 \label{FIG_3}
\end{figure*}

\subsection{Traditional normal estimation}
The most famous normal estimator is principal component analysis (PCA)\cite{hoppe1992surface}. This kind of regression-based method samples fixed neighbors and fits a local tangent plane. Several variants such as local spherical surfaces fitting\cite{guennebaud2007algebraic} and Jets\cite{cazals2005estimating} are proposed to fit the higher-order local surface and improve the normal accuracy effectively. However, the performance of these approaches usually depends on the scale of the selected patches and the noise level. The key to solving the problem is how to choose reasonable candidate points for a given patch with different sampling sizes and noise scales. Mitra et al.\cite{mitra2003estimating} adapt the sampling radius by investigating the effect of local noise, curvature, and sampling density. Pauly et al.\cite{pauly2003shape} assign a Gaussian weight to each neighbor. Nevertheless, the problem of isotropic sampling is still not solved. More robust statistics approaches are employed to select a plane approximating the neighbors from the same surface patch as the current point\cite{li2010robust,fleishman2005robust,yoon2007surface,mederos2003robust,wang2013adaptive,wang2013consolidation}. While many of these methods hold theoretical guarantees on approximation and robustness, all the above methods require a careful setting of parameters. So the data-driven methods are urgently necessary technologies for more robust to estimate the normals of 3D models. 

\subsection{Learning-based normal estimation}
As a pioneer, Boulch and Marlet \cite{boulch2016deep} first apply 2D convolutional neural network (CNN) to regress the point normal. A transformed hough space accumulator is used as the input of a CNN to fit the local plane. However, this method is based on image input and does not use the point data directly. Inspired by PointNet\cite{qi2017pointnet}, PCPNet\cite{guerrero2018pcpnet} is proposed to regress the normal based on local neighbors. Besides, based on 3D modified fisher vectors (3DmFV) presentations, Nesti-Net\cite{ben2019nesti} is proposed to use a mixture-of-experts architecture to select the best scale for each point, but it caused a great time consumption. Then, Zhou et al. \cite{zhou2020normal} introduce an extra plane feature constraint mechanism as a regularization item to improve the robustness of the regressor. Considering the classic problem-specific knowledge strategy, Cao et al. \cite{9339852} propose a differentiable RANSAC-like module to predict a latent tangent plane, and Lenssen et al. \cite{lenssen2020deep} use a learnable anisotropic kernel to fit the local tangent plane by an iteration way, denoted by IterNet. Although this kind of strategy can effectively improve the performance of a learning-based regressor, the simple local 1-order plane hypothesis is not suitable for rich patch styles. Thus, DeepFit\cite{ben2020deepfit} incorporates a neural network to learn point-wise weights for weighted least squares polynomial surface fitting. However, we find that the learning-based networks are not well compatible with the classic surface fitting method, as shown in Fig. \ref{FIG_1}. In this paper, two simple-yet-effective strategies (see Fig. \ref{FIG_2}) are proposed to enhance the compatibility between the traditional fitting method and the weight learned network.

\section{Overview}
This paper presents two strategies that can be inserted between the weight-learning neural network and fitting-based normal estimation architecture, which can dramatically improve the normal estimation result (See Fig. \ref{FIG_3}). Similar to DeepFit \cite{ben2020deepfit}, given a local patch, we first use a neural network to estimate the weight of each point, but instead of directly introducing a weighted least squares surface fitting method, we use a top-k selection strategy (Sec. \ref{sec:4-3}) to effectively evaluate the most critical points in a patch at the training stage. Besides, we introduce a point update strategy (Sec. \ref{sec:4-4}), which can update the point position based on a geometric guidance module. The updated points will be used in the surface fitting process, which dramatically relieves the distortion situation (see the first row of Fig. \ref{FIG_1}), thus producing a reasonable and simplified surface as shown in the third row of Fig. \ref{FIG_1} with the order $n>2$. Finally, the Jet can estimate the center point's normal more precisely based on the selected points with position update.

\section{Approach}
\subsection{Pre-processing}
Firstly, given a 3D point cloud $X = \{p_{1},p_{2},\cdots,p_{N}\} \in R^{N \times 3}$,  we can sample a local patch $P_{i} = \{p_{i,j} | p_{i,j} \in  KNN(p_{i})\}$ for each point $p_{i}$, which is constructed by the K nearest neighbor (kNN) search. Here, $j=1,2,\cdots,r$ and $r$ is the number of sampling scale for a given patch. Then, for each patch, a classic PCA method is used to align the patch as the initial input of our patch-based network.

\subsection{Network Architecture}
The feature learning network is similar to PCPNet \cite{guerrero2018pcpnet}, which consists of two transformation layers and a multi-layer perceptrons (MLPs) with four layers. The network can output point-wise features, which can be used to estimate the weight of each point from the patch. Here, we also use a quaternion spatial transformer (QST) to transfer the input patch to a canonical pose. Here a subnetwork is used to evaluate the quaternion that parameters a local spatial transform. The output of quaternion can be converted to a rotation matrix. Then the rotated patch is feed into the normal estimation network. A significant property of the network is that it should be invariant to the input point ordering.

As shown in Fig. \ref{FIG_3}, for an aligned patch $P_{i}=$ $\{ p_{i,1}, p_{i,2}, \cdots, p_{i,r} \}$, our feature learning network outputs point-wise feature $F_{i} = \{f_{i,1},f_{i,2},\cdots,f_{i,r}\}$. We use the simple Max-Pooling operator to aggregate the features of the input patch to output a global point cloud representation denoted as $G_{i} = Max \{F_{i}\}$. The point-wise features and the global representation are then concatenated and fed into a multi-layer perceptron $h(\cdot)$ followed by a sigmoid activation function. Finally, the output $W_{i} = \{w_{i,1},w_{i,2},\cdots,w_{i,r}\} \in R^{r} $ of this network is a weight per point for the given patch $P_{i}$:
\begin{equation}
w_{i,j} = sigmoid(h(f_{i,j} \bigoplus G_{i}))
\end{equation}
where $j=1,2,\cdots,r$ and $\bigoplus$ is a concatenation operator. The learned weights will be used to fit local patch and estimate the center normal of the patch.

\begin{figure}[t]
 \centering 
 \includegraphics[width=8cm]{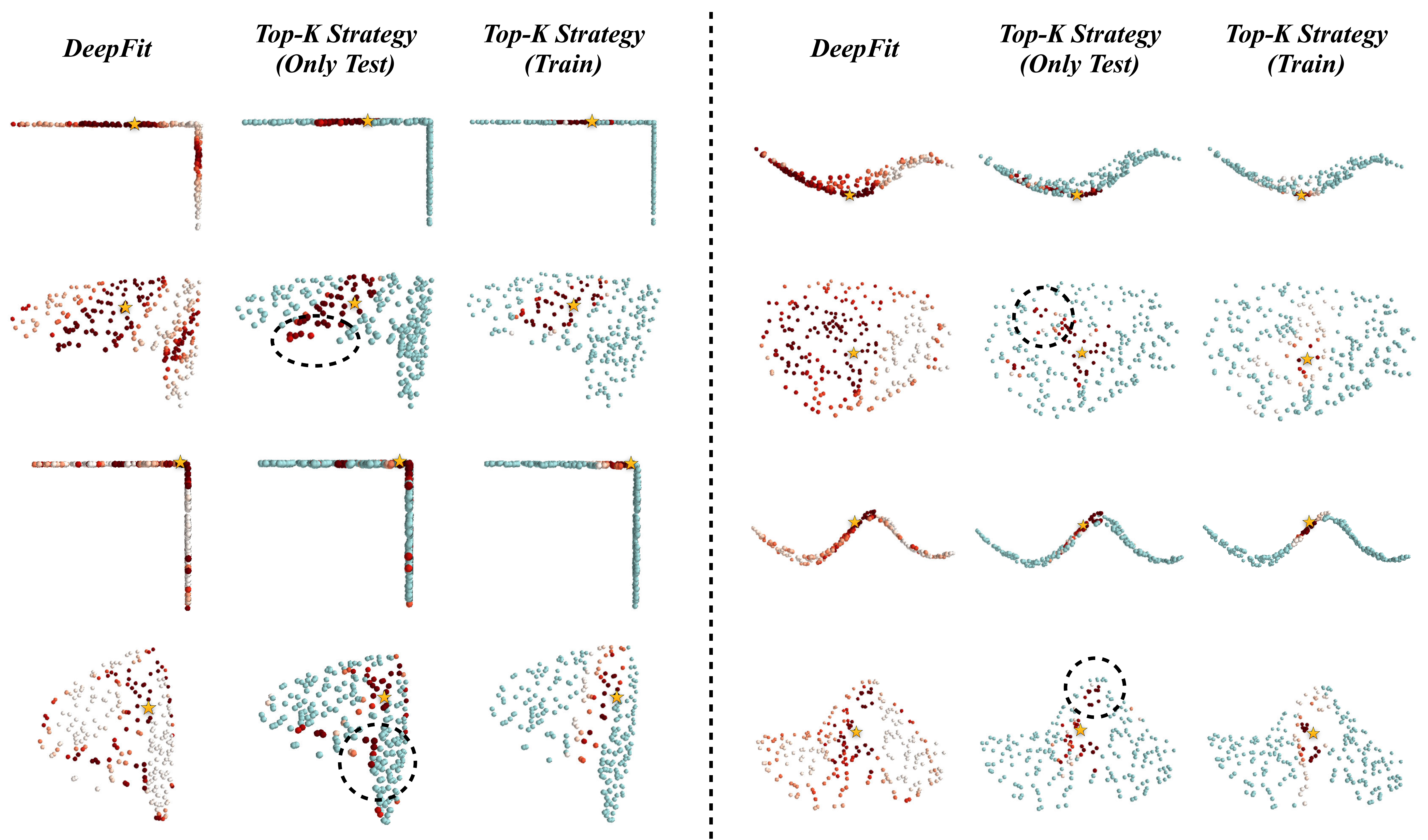}
 \caption{Comparisons of the visual weights learned from DeepFit\cite{ben2020deepfit} and our top-k selection strategy. Each group from left to right: the weight leaned from DeepFit, the top-k weights selection without training process and the trained top-k selection strategy. We obtain a more reasonable and stable weight map on a given patch.}\label{FIG_4}
\end{figure}

\subsection{Top-k selection strategy}\label{sec:4-3}
We present a top-k selection strategy in the weight learning network to focus on the most critical points for normal estimation task. It can be seen as a simplification process: the local points that belong to the underlying plane where the center point is located are the critical points for fitting, and a more simple surface fitting process will obtain better estimation results, as shown in the fourth row of Fig. \ref{FIG_1}. 

Therefore, during training, we introduce a simple strategy between the weight learning network and the surface fitting algorithm to learn more stable weight for each point without the ambiguous weight redundancy, which can be observed in the first and second rows of each group in Fig. \ref{FIG_4}). In our architecture, the estimated point-wise weights are first fed into the top-k selection module, and then we dynamically sort the learned weights in a descending order based on the value and preserve the first k points and weights for the normal estimation by a more simplified fitting process in last. Benefit from the dynamic selection of the first k points during training, and the backward process can better force the network to focus on the most critical points and evaluate the importance of the patch points more exactly. Specifically, the pipeline of the top-k selection process is illustrated in Fig. \ref{FIG_3}. Given the input patch $P_{i} \in R^{r\times 3}$ and the learned weights $W_{i}$, then we sort the weights and select the first top k values 
denoted as $\widetilde{W}_{i} \in R^{k}$. The indices of the k selected value are also kept.


Figure \ref{FIG_4} shows several examples, compared to the weights learned by DeepFit \cite{ben2020deepfit} (the first column of each example), the top-k learning process can help the network to obtain a more reasonable weights map which is helpful to normal estimation. Besides, the top-k selection strategy must be used at both the learning and inference stages, which can effectively improve the network to focus on the critical points of the patch. Without the dynamic top-k selection learning, just a simple top-k selection at the inference stage is not helpful, as shown in the middle column of each group in Fig. \ref{FIG_4}.

\begin{figure}[t]
 \centering 
 \includegraphics[width= 0.94\columnwidth]{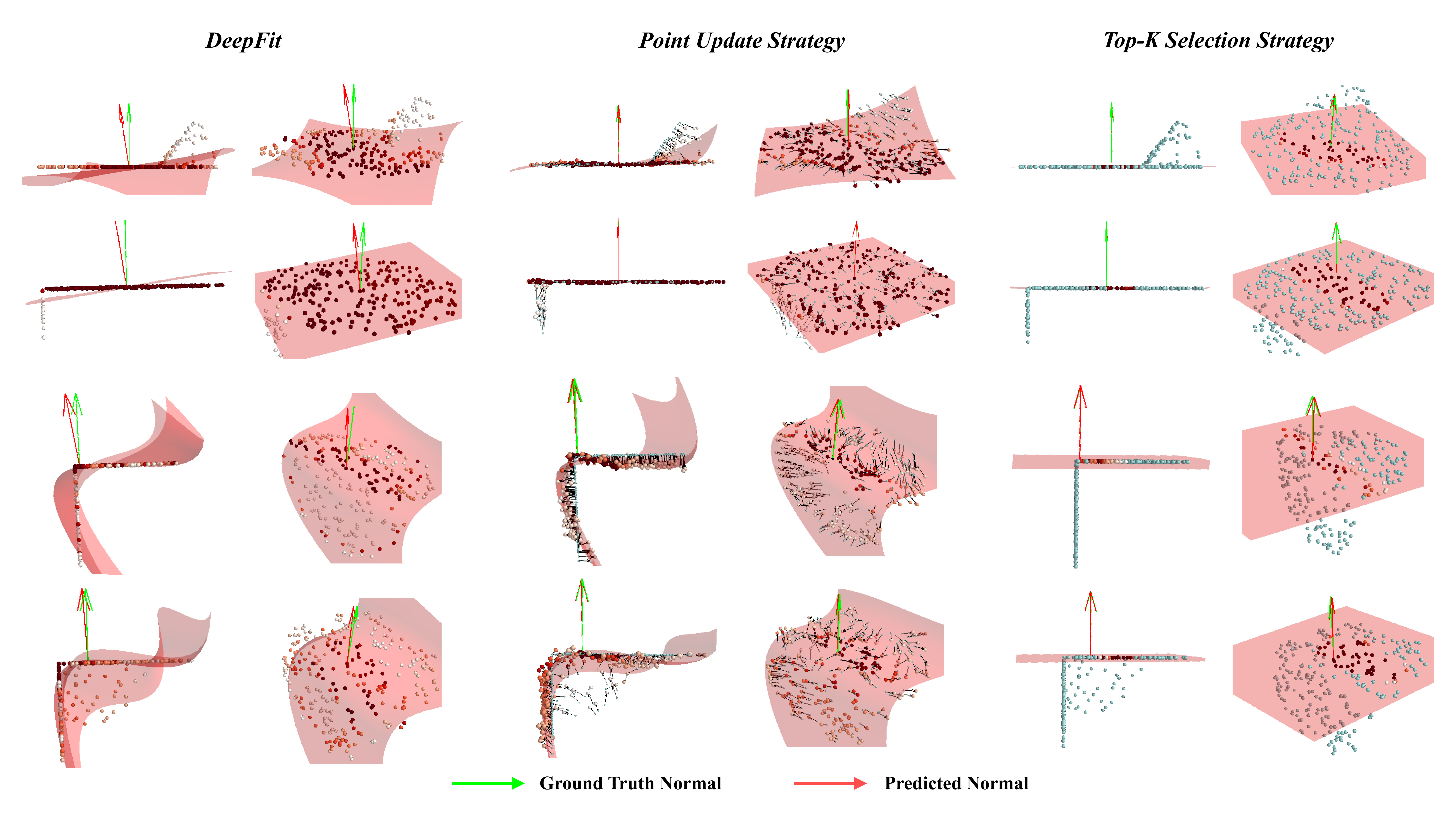}
 \caption{Surface fitting and normal estimation visualization results of our method compared to DeepFit\cite{ben2020deepfit}. Compared with Deepfit\cite{ben2020deepfit}, our proposed strategies ensure that a more concise surface is fitted near the sharp boundary, which also promotes a more accurate normal to be predicted.}
 \label{FIG_5}
\end{figure}

\subsection{Point-wise position update strategy}\label{sec:4-4}
In this paper, we also present a valuable strategy to improve the fitting process to overcome the fitting distortion, especially near the sharp edge of a patch, which will significantly affect the center point's normal estimation. This situation is shown in the first and second rows of Fig. \ref{FIG_1}. Thus, we introduce a simple deformation sub-network to update the position of each point to produce a more smooth patch. The 2D illustration of this operation is given in the second column of Fig. \ref{FIG_2}. We can see that the deformation patch can be fitted using a more simple parameter surface and preserving a consistent local structure near the patch center to ensure the accuracy of the normal estimation. 


Specifically, for each point $p_{i,j}$ on the given patch $P_{i}$, we first search its $m$ nearest neighbors $\{p_{i,j,1},p_{i,j,2}, \cdots,p_{i,j,m}\} \in P_{i}$. Then, for each point of the patch, the local edge vectors can be denoted as $p_{i,j,s}-p_{i,j}$,  we utilize the local edge vector as a guidance during learning. As shown in the Point-wise Position Update parts of Fig. \ref{FIG_3}, our module transforms semantic information of each edge into its weight and then we aggregate the weighted edge vectors together to obtain our predicted offset direction. Intuitively, the prediction shift is decided by the voting of surrounding edges with different significance. For the $j$-th point of a patch, its offset can be computed as:
\begin{equation}
\Delta p_{i,j} = \frac{1}{m}\sum_{s=1}^{m} R(f(p_{i,j,s}) - f(p_{i,j}))\cdot(p_{i,j,s}-p_{i,j})
\end{equation}
where $p_{i,j}$ and $p_{i,j,s}$ denote the location of a point and its neighbors of the input patch $P_{i}$, so $p_{i,j,s}-p_{i,j}$ means the edge direction. $f(p_{i,j,s})$ and $f(p_{i,j})$
are their learned  point feature extracted from the the PointNet architecture. $R(\cdot)$ can obtain the weight from one convolution layer via transforming point features. After obtaining learned shift $\Delta p_{i,j}$, we achieve a new patch with the self-adaptive update:
\begin{equation}
p_{i,j}^{'} = p_{i,j}+\Delta p_{i,j}
\end{equation}
Finally based on the top-k selected indices we can obtain a local patch $P_{i}^{'} \in R^{k \times 3}$ for final surface fitting and normal estimation.

\subsection{Jet fitting}
Given a local patch with PCA alignment, the neural networks dynamically select the patch's top-k critical points and adjust the points to a suitable position for surface fitting. Then, similar to DeepFit \cite{ben2020deepfit}, we introduced truncated Taylor expansion surface fitting \cite{cazals2005estimating} using weighted least-squares (WLS). We refer to the truncated Taylor expansion as a degree n jet or n-jet for short. This surface embedding process can be written as a "height function":
\begin{equation}
f(x,y) = J_{\beta,n}(x,y) = \sum_{s=0}^{n}\sum_{t=0}^{s}\beta_{s-t,t} x^{s-t} y^{t}
\end{equation}
where $\beta$ is  the jet coefficients vector that consists of $N_{n} = (n+1)(n+2)/2$ terms. 

Given the selected and updated points $P_{i}^{'}$ including k points, we first specify a Vandermonde matrix $ M = (1,x_{j},y_{j},\cdots, x_{j}y_{j}^{n-1},y_{j}^{n})_{j=1,2,\cdots,k} \in R^{k\times N_{n}}$ and the height vector can be denoted as $B = (z_{1},z_{2},\cdots,z_{k})^{T} \in R^{k}$ representing the sampled points. Then, based on the n-jet height function over a surface, we can obtain a system of linear equation based on the points:
\begin{equation}
M \beta = B
\end{equation}
When $N_{n} < k$, the system is over-determined and an exact solution may not exist. Therefore we use an LS approximation that minimizes the sum of square errors between the value of the jet and the height function over all points:
\begin{equation}\label{eq:6}
\beta = \arg \min_{z\in R^{N_{n}}}\|Mz-B\|^{2}
\end{equation}

Generally, the sampled patch includes noise and outliers, which heavily reduce the fitting accuracy. To overcome this,  we use the formulation given in Eq. \ref{eq:6}, which is a weighted least square problem. Given the weights of the selected top-k points, a diagonal weight matrix can be constructed $W = diag(w_{1},w_{2},\cdots,w_{k})$. Then, the optimization problem becomes:
\begin{equation}
\beta = \arg \min_{z\in R^{N_{n}}}\|W^{1/2}(Mz-B)\|^{2}
\end{equation}
Finally, we can obtain the solution:
\begin{equation}
\beta = (M^{T}WM)^{-1}M^{T}WB
\end{equation}
The estimated normal of center point is given by:
\begin{equation}
N_{i} = \frac{(-\beta_{1},-\beta_{2},1)}{\|(-\beta_{1},-\beta_{2},1)\|_{2}}
\end{equation}
Then the neighbors' normals of the selected points can also be calculated by:
\begin{equation}
N_{j} = \frac{\nabla F}{\|\nabla F\|}|_{p_{i,j}^{'}} = \frac{(-\beta \frac{\partial M^{T}}{\partial x},\beta \frac{\partial M^{T}}{\partial y},1)}{\|\nabla F\|} \vert_{p_{i,j}^{'}}
\end{equation} 

\begin{figure*}[htp]
 \centering 
 \includegraphics[width= 0.9\textwidth]{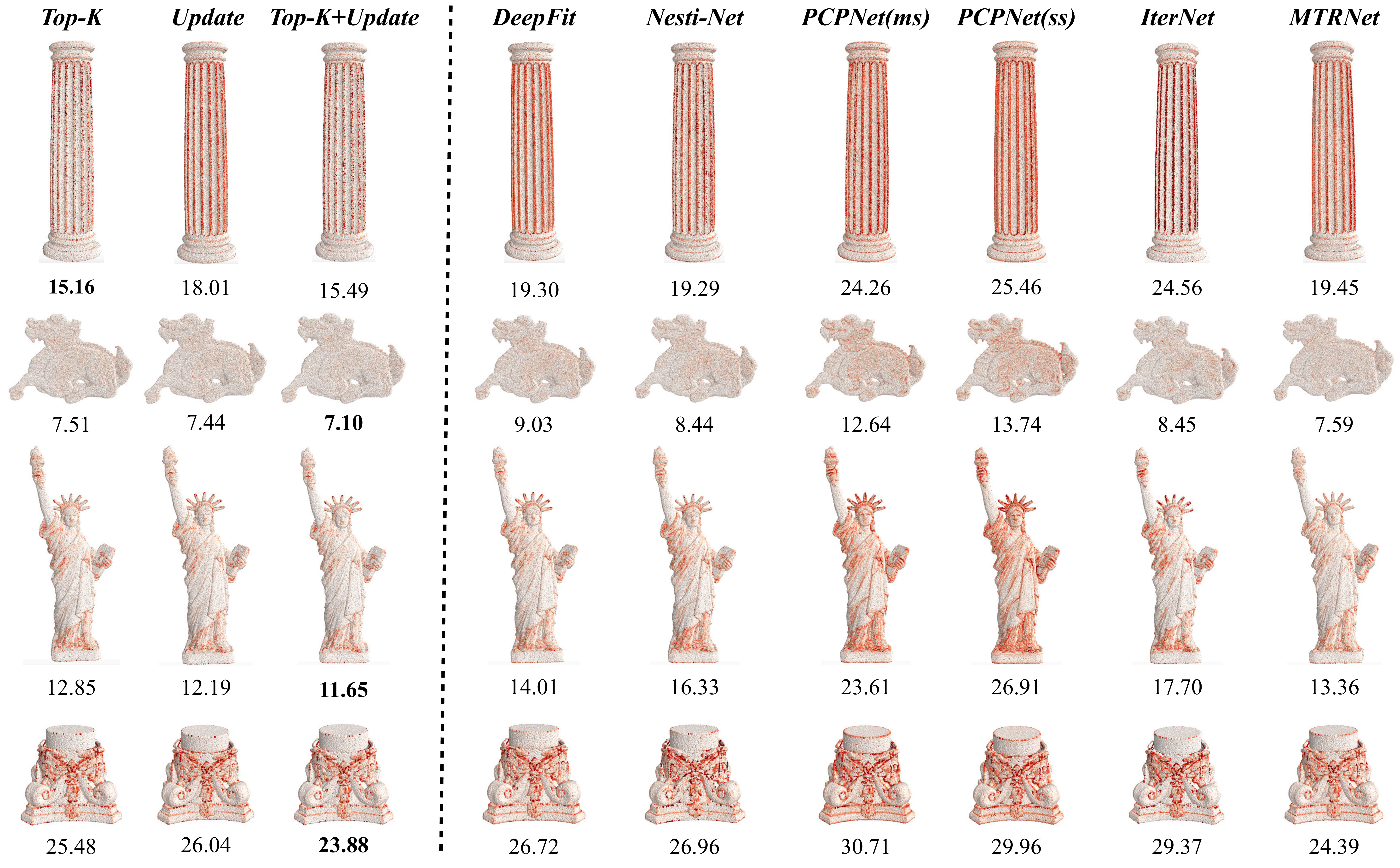}
 \caption{Angular error visualization results of our method compared to others. The colors of the points correspond to angular difference, mapped to a heatmap ranging from 0-90 degrees. }
 \label{FIG_6}
\end{figure*}

\subsection{Loss Function}
Note that this formulation assumes all points to lie on the surface. For points that are not on the surface, the normal error will be large. Therefore that points weight will be encouraged to be small. This term can easily converge to an undesired local minimum by setting all weights to zero. To avoid that, we add a regularization term that computes the negative average log of the $k$ selected weights. In summary, the consistency loss is then given by:
\begin{equation}
L_{neighbors} = \frac{1}{k}[-\sum_{j=1}^{k}\log(\widetilde{w}_{i,j})+\sum_{j=1}^{k}\widetilde{w}_{i,j}|N_{gt}\times N_{j}|]
\end{equation}
Then Similar to PCPNet, for the given patch $P_{i}$, the normal prediction loss performed by minimizing the sin loss between estimated normal $\hat{N}_{i}$ and ground truth $N_{GT}$ as usual:
\begin{equation}
L_{center} = |N_{gt}\times \hat{N}_{i}|
\end{equation}

Then, with the PointNet' s transformation matrix regularization terms $L_{reg} = |I-AA^{T}|$, our final training loss is:
\begin{equation}
L_{total} = L_{center} +\alpha_{1}L_{neighbors}+\alpha_{2}L_{reg}
\end{equation}
Here, $\alpha_{1}$, and $\alpha_{2}$ are weighting factors, chosen empirically

\begin{table*}[tb]
\caption{Comparison of the RMSE angle error for unoriented normal estimation of our method (two sampling sizes) to classical geometric methods, and deep learning methods.}
  \scriptsize%
	\centering%
\begin{tabular}{l ccc ccc ccc c cc ccc}
\toprule
\multirow{2}{*}{} & \multicolumn{2}{c}{\textbf{Ours}}&\multicolumn{3}{c}{\textbf{PCA}} &\multicolumn{3}{c}{\textbf{Jet}}& \multicolumn{1}{c}{\textbf{HoughCNN}}&\multicolumn{2}{c}{\textbf{PCPNET}} & \textbf{Nesti-Net}&\textbf{Lenssen}&\textbf{DeepFit}\\
\cmidrule(r){2-3} \cmidrule(r){4-6}\cmidrule(r){7-9}\cmidrule(r){10-10} \cmidrule(r){11-12} \cmidrule(r){13-13}\cmidrule(r){14-14}\cmidrule(r){15-15}
&256 &512 &small &med &large
&small &med &large
&ss &ss &ms &ms(MoE) &ss &ss \\
\midrule
None         &\textbf{5.33}   &5.90   &8.31  &12.29 &16.77  &7.60  &12.35 &17.35  &10.23  &9.68  &9.62   &6.99   &6.72   &6.51\\
\hline
\textbf{Noise} $\sigma$ \\
0.00125      &9.21   &\textbf{9.10}   &12.00 &12.87 &16.87  &12.36 &12.84 &17.42  &11.62  &11.46 &11.37  &10.11  &9.95   &9.21 \\
0.006        &16.95  &\textbf{16.50}  &40.36 &18.38 &18.94  &41.39 &18.33 &18.85  &22.66  &18.26 &18.37  &17.63  &17.18  &16.72\\
0.012        &23.19  &22.08 &52.63 &27.50 &23.50  &53.21 &27.68 &23.41  &33.39  &22.80 &23.28  &22.28  &\textbf{21.96}  &23.12\\
\hline
\textbf{Density} \\
Gradient     &\textbf{6.16}   &6.40   &9.14  &12.81 &17.26  &8.49  &13.13 &17.80  &12.47  &13.42 &11.70  &9.00   &7.73   &7.31\\
Stripes      &\textbf{6.28}  &6.79   &9.42  &13.66 &19.87  &8.61  &13.39 &19.29  &11.02  &11.74 &11.16  &8.47   &7.51   &7.92\\
\hline
Average     &11.18  &\textbf{11.13}  &21.97 &16.25 &18.87  &21.95 &16.29 &19.02  &16.90  &14.56 &14.34  &12.41  &11.84  &11.80\\
\bottomrule
\end{tabular}
\label{tab:1}
\end{table*}

\section{Experiments}
\subsection{Dataset and training details}
For training and testing, we use the PCPNet dataset\cite{guerrero2018pcpnet}, which includes both CAD objects and high-quality scanned models. In total, it consists of 8 models for training and 22 models for testing. All shapes are given as triangle meshes and densely sampled with 100k points. The data is augmented by introducing i.i.d. Gaussian noise for each point's spatial location with a standard deviation of 0.012, 0.006, 0.00125 w.r.t the bounding box size. For evaluation, we use the same 5000 point subset per shape as in Guerrero et al. \cite{guerrero2018pcpnet}. The batch size and the base learning rate are set to 48 and 0.1, respectively.  The Adam optimizer is used to train our network. The implementation was done in PyTorch and trained on a single Nvidia RTX 2080 Ti GPU.

\begin{table*}[h]
\caption{The comparisons of RMSE angle error with diff. sampling sizes and diff. value K for the Top-K Selection Strategy.}
\centering
\begin{tabular}{l  ccccc  cccc}
\toprule
\multirow{2}{*}{} & \multicolumn{5}{c}{\textbf{256 points}}&\multicolumn{4}{c}{\textbf{512 points}} \\
\cmidrule(r){2-6} \cmidrule(r){7-10}
\textbf{Top $k$} &30      &50   &100   &150 &256    &50& 100 &150  &512\\
\midrule
None    &6.45  &5.48   &5.73      &5.80	&6.51         &6.19 &6.07	&6.27 & 7.19\\
\hline
\textbf{Noise $\sigma$ }  \\
0.00125 &9.63  &9.19   &9.02     &8.98	&9.21             &9.31 	&9.18	&9.19 &9.40\\
0.006   &16.91 &16.87  &16.79   &16.80	&16.72         &16.71	&16.57	&16.51 &16.25\\
0.012   &22.63 &23.29  &23.08    &23.18	&23.12       &22.30 	&22.12	&22.13 &21.77\\
\hline
\textbf{Density}\\
Gradient  &6.76 &6.27   &6.52      &6.58	&7.31               &6.58 	&6.45	&6.68 &7.86\\
Stripes   &7.24   &6.38   &6.78     &6.96	&7.92              &7.00	&6.86	&7.00 &8.80 \\
\hline
\textbf{average}  &11.60 &\textbf{11.25}  &11.32   &11.38	&11.8  &11.35 	&\textbf{11.21}	&11.30 &11.88\\ 
\bottomrule
\end{tabular}
\label{tab:4}
\end{table*}

\begin{table*}[h]
\caption{Effects of the proposed strategies: The top-k selection strategy and point update strategy. The RMSE angle error is used on PCPNet dataset. The sampling sizes are set to 256 and 512 points, respectively.} 
\scriptsize
\centering
\begin{tabular}{l  cccc  cccc}
\toprule
\multirow{2}{*}{} & \multicolumn{4}{c}{\textbf{256 points}}&\multicolumn{4}{c}{\textbf{512 points}} \\
\cmidrule(r){2-5} \cmidrule(r){6-9}
 \textbf{Scale} &\textbf{Baseline} &\textbf{ \tabincell{c}{Ours\\(update)}} &\textbf{\tabincell{c}{Ours\\(top-K)} }&\textbf{\tabincell{c}{Ours\\(update\\+top-k)}}&\textbf{Baseline} &\textbf{ \tabincell{c}{Ours\\(update)}} &\textbf{\tabincell{c}{Ours\\(top-K)} }&\textbf{\tabincell{c}{Ours\\(update\\+top-k)}}\\
\midrule
None        &6.51   &5.86   &5.48   &5.33	&7.19	&6.33	&6.07	&5.90\\
\hline
\textbf{Noise $\sigma$ }                          \\
0.00125     &9.21  & 9.19  &9.19  &9.21	&9.40	&9.31	&9.18	&9.10\\
0.006        &16.72  &16.73  &16.87  &16.95	&16.25	&16.37	&16.57	&16.50\\
0.012       &23.12 &22.90  &23.29  &23.19		&21.77	&21.85	&22.12	&22.08\\
\hline
\textbf{Density}\\
Gradient    &7.31  &6.64   &6.27   &6.16	&7.86	&6.76	&6.45	&6.40\\
Stripes     &7.92  &6.94  &6.38  &6.29	&8.80	&7.56	&6.86	&6.79\\
\hline
\textbf{average} &11.80     &11.37 &11.25 &\textbf{11.18}	&11.88	&11.36	&11.21	&\textbf{11.13}\\
\textbf{improv.}  &- &3.64\% &4.66\%&5.25\%	&- &4.38\%	&5.64\%	&6.31\%\\

\bottomrule
\end{tabular}

\label{tab:5}
\end{table*}

\subsection{Normal estimation performance}
RMSE angle error of our approaches and related methods on the PCPNet test set are shown in Tab. \ref{tab:1}. For HoughCNN \cite{boulch2016deep}, we use the single-scale network provided by the authors. For PCPNet \cite{guerrero2018pcpnet}, we use single-scale and multi-scale networks provided by the author, compared with classification methods. Nest-Net \cite{ben2019nesti}  achieves higher accuracy compared to PCPNet \cite{guerrero2018pcpnet}. However, it is an MoE architecture with seven sub-networks using a more significant number of parameters than other methods. DeepFit \cite{ben2020deepfit}, considering the local latent surface representation, there is a great improvement. IterNet \cite{lenssen2020deep} iterates to estimate point normal and have lower RMS errors than the above methods, but the network needs to input the whole model simultaneously, which is not flexible. It can be seen that our method can achieve better performance with two sampling scales. Specifically, the large sampling scale will achieve better results for the models with large noise and small scale is more beneficial to low noise data. Besides, Fig. \ref{FIG_6} also depicts the angular error in each point for the different learning-based methods using a heat map. It is seen that the methods (both using the top-k selection and point update strategies) used in this paper have consistent advantages compared to other methods for both models with smooth details and sharp features.

\begin{figure}[htp]
 \centering 
 \includegraphics[width=1\columnwidth]{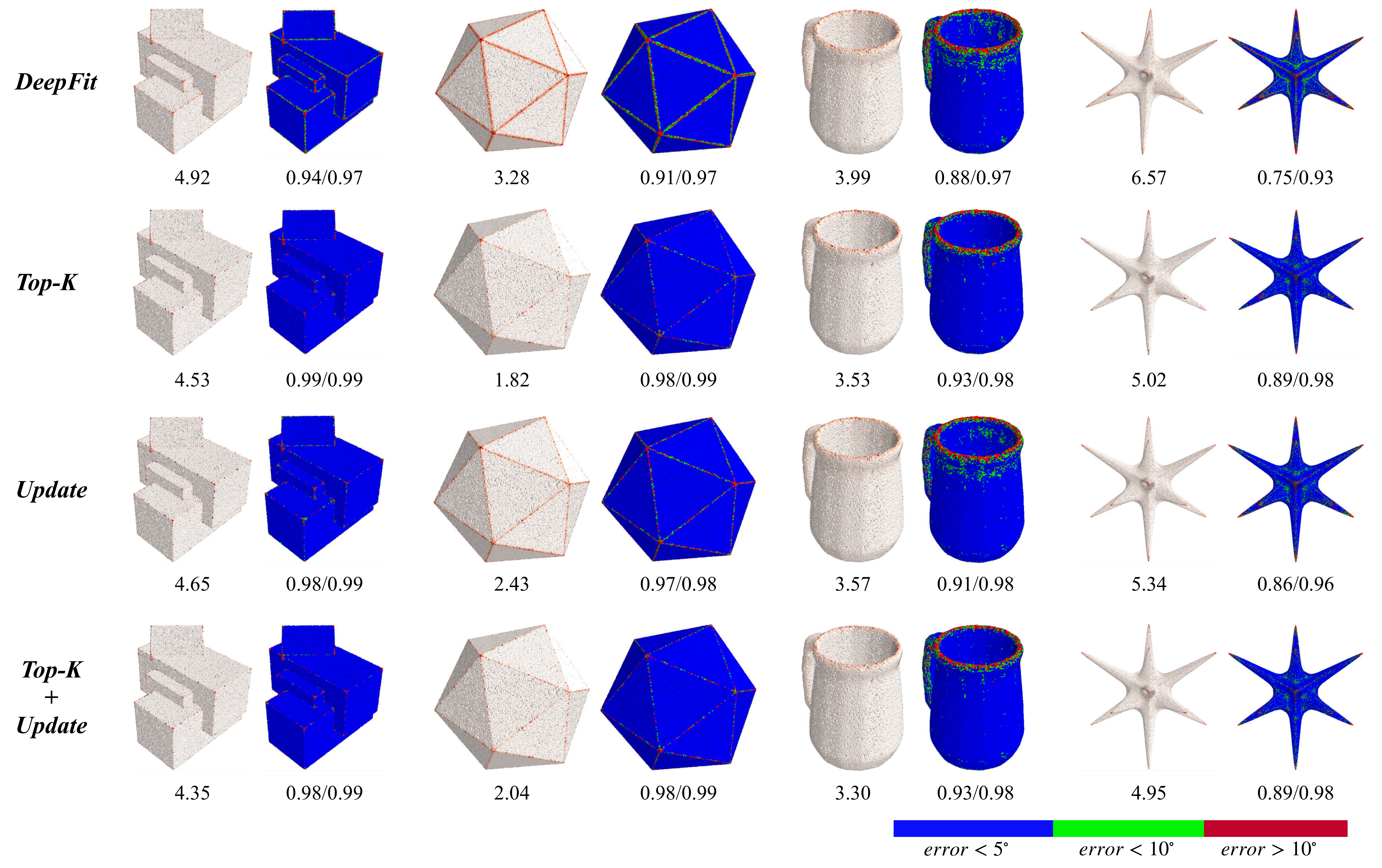}
 \caption{Quantitative analysis of the normal estimation. Blue denotes points with an estimated normal that deviates by less than $5^{\circ}$ from the ground-truth; green encodes angular deviations between $5^{\circ}$ and $10^{\circ}$ ; red marks errors $>10^{\circ}$.}
 \label{FIG_7}
\end{figure}

\subsection{The Visualization of Surface Fitting Process}
We also give four visual examples about the surface fitting results using the different strategies shown in Fig. \ref{FIG_5}. The point update strategy can dramatically reduce the distortion near the sharp edge for the fitted surface. In practically, for the second row of Fig. \ref{FIG_5}, the point update module can push the points near the boundary to a consistent plane part, and the fitting process can be simplified at last. Besides, near the corner of a patch, as shown in the last row of Fig. \ref{FIG_5}, the points near the corner can be pushed to a more smooth pattern, and the updated patch is more likely to be reconstructed by a surface with low order.

Additionally, the third column of Fig. \ref{FIG_5} also shows the visualizations of surface fitting results by using the top-k selection strategy. Benefited from dynamical weights learning from our network (as shown in Fig. \ref{FIG_4}, we can obtain more reasonable weights for describing critical points of a given patch), we can make the surface fitting process focus on the more critical plane in a patch. Ignoring the points on other planes can simplify the fitting process and improve the accuracy of normal estimation.

\subsection{Ablation Study}
\noindent
\textbf{The performance of point update strategy.} 
Typically, the sampled patch includes sharp feature that heavily effects the fitting process and the normal accuracy. Thus, we consider to learn to adjust the positions of the points from a patch to more suitable to surface fitting while preserve the accurate normal of the center point. Tab. \ref{tab:5} shows the improvements in two scale sampling. For 256 and 512 points sampled, this strategy can gives a 3.64 $\%$ and 4.38 $\%$ improvements respectively.

\noindent
\textbf{The performance of top-k selection strategy.} 
Generally, a sampled patch may contains multiple surfaces, our goal is to estimate a correct normal from the main plane which the center point belonging to. In this paper we introduce a simple top-k selection in both training and interface stages to dynamically learn the critical points, which can dramatically simplify the fitting process and make our algorithm focus most on the main plane points as shown in Fig. \ref{FIG_4} and Fig. \ref{FIG_1}. These means that a simple 1-order surface is always enough for the normal estimation task. Also, see the Tab.~\ref{tab:5},  the top-k selection module gives a 4.66$\%$ and 5.64$\%$ performance boost compared to DeepFit \cite{ben2020deepfit} for 256 and 512 points sampling respectively.

\noindent
\textbf{The visualizations PGP$\alpha$.} For more clearly to seen the improvements of our proposed strategies in this paper, we also  use the proportion of good points metric (PGP$\alpha$), which computes the percentage of points with an error less than $\alpha$; e.g., PGP10 computes the percentage of points with angular error of less than 10 degrees. By using the top-k selection strategy, the estimated normals can by more accurate near the sharp boundary. The final RMSE and PGP5/10 are also given in the bottom of each model. For the models with sharp features or smooth details, our strategies can give noticeable improvement compared to the method of DeepFit \cite{ben2020deepfit}.

\noindent
\textbf{The selection of $k$ and sampling size.}
The RMSE error of different sampling sizes and parameter $k$ in top-k module are explored (see Tab.~\ref{tab:4}). Firstly, it shows that a large sampling size will improve the high noise cases, and for large noise case, we need to use a large $k$ to ensure the accuracy. The table also shows that both the small and large k will improve the accuracy of the normal estimation compared to the results without the top-k strategy. We can choose $k=50$ for 256 points sampled and $k=100$ for 512 points sampled finally.

\section{Conclusion}
Two simple and effective strategies are proposed to enhance the compatibility of the learning-based method and classic surface fitting algorithm, which can improve the accuracy of normal estimation for the point cloud. Firstly, the top-k selection strategy is employed to dynamically select the critical points of a patch and make the surface fitting process focus on the essential part, which can dramatically improve the point's prediction accuracy. Then, the point update strategy can deform the sharp corners of a patch and alleviate the difficulty of local surface fitting and reduce the distortion of the fitting surface, improving the precision of normal estimation. The approach demonstrates the competitiveness compared to SOTA approaches.


\bibliographystyle{Bibliography/IEEEtranTIE}
\bibliography{Bibliography/IEEEabrv,Bibliography/BIB_xx-TIE-xxxx}\ 

\end{document}